\documentclass[10pt,twocolumn,letterpaper]{article}

\usepackage{cvpr}
\usepackage{times}
\usepackage{epsfig}
\usepackage{graphicx}
\usepackage{amsmath}
\usepackage{amssymb}
\usepackage[normalem]{ulem}
\usepackage{subfig}
\usepackage{caption} 
\newcommand{\norm}[1]{\lvert\lvert #1 \rvert\rvert}

\usepackage[pagebackref=true,breaklinks=true,letterpaper=true,colorlinks,bookmarks=false]{hyperref}

 \cvprfinalcopy 

\def\cvprPaperID{(Left blank intentionally)} 

\ifcvprfinal\fi

\usepackage{color}

\begin{document}

\title{Aerial Vehicle Tracking by Adaptive Fusion of Hyperspectral Likelihood Maps}

\author{Burak Uzkent\\
Chester F. Carlson Center for Imaging Science\\
Rochester Institute of Technology\\
{\tt\small bxu2522@@rit.edu}
\and
Aneesh Rangnekar\\
Chester F. Carlson Center for Imaging Science\\
Rochester Institute of Technology\\
{\tt\small apr2635@@rit.edu}
\and
Matthew J. Hoffman\\
School of Mathematical Sciences\\
Rochester Institute of Technology\\
{\tt\small mjhsma@rit.edu}
}

\maketitle

\begin{abstract}
Hyperspectral cameras provide unique spectral signatures that can be used to solve surveillance tasks. This paper proposes a novel real-time hyperspectral likelihood maps-aided tracking method (HLT) inspired by an adaptive hyperspectral sensor. We focus on the target detection part of a tracking system and remove the necessity to build any offline classifiers and tune large amount of hyper-parameters, instead learning a generative target model in an online manner for hyperspectral channels ranging from visible to infrared wavelengths. The key idea is that our adaptive fusion method can combine likelihood maps from multiple bands of hyperspectral imagery into one single more distinctive representation increasing the margin between mean value of foreground and background pixels in the fused map. Experimental results show that the HLT not only outperforms all established fusion methods but is on par with the current state-of-the-art hyperspectral target tracking frameworks.
\end{abstract}

\section{Introduction}
Vehicle detection and persistent tracking in aerial imagery is an extremely challenging problem due to the need for accurate movement tracking of vehicles under all possible circumstances - traffic, low resolution due to platform height, camera motion compensation, parallax, occlusion, environmental interference, and low frame rate to name a few. In spite of these challenges, aerial vehicle tracking has attracted considerable interest over the past few years due to its growing importance in various applications. 

\begin{figure}[t]
\begin{center}
   \includegraphics[width=1\linewidth]{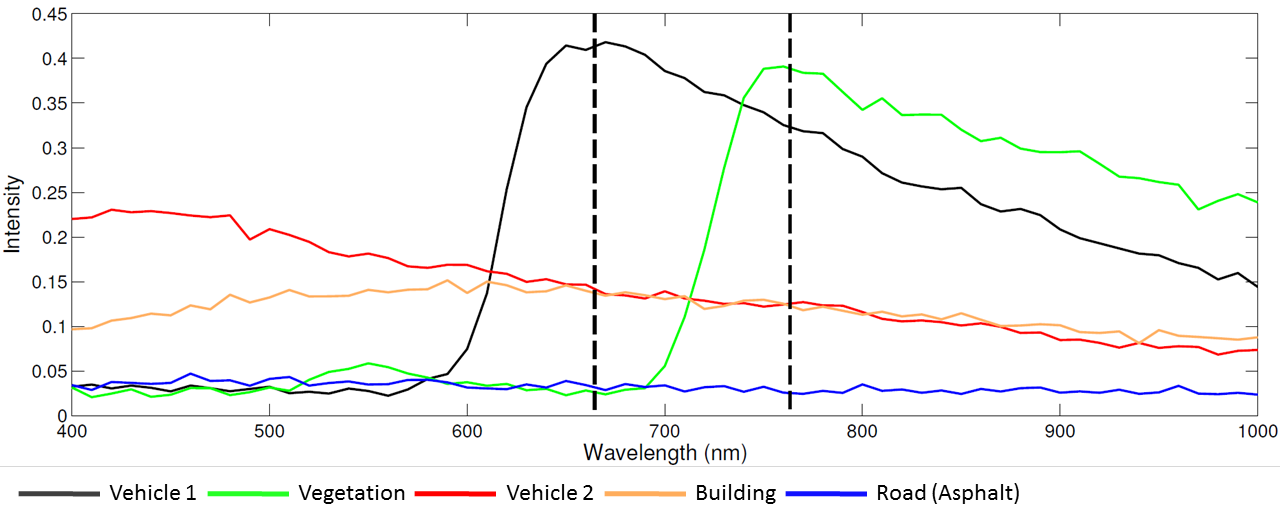}
\end{center}
   \caption{Spectral reflectances of different objects commonly found in an urban scene. As observed, vehicles 1 and 2 (Black and Red respectively) can be distinguished between themselves and from background objects if hyperspectral signatures are put to use.}
\label{fig:specprofiles}
\end{figure}

The ultimate goal of an aerial vehicle tracking system is to continuously track target(s) of interest through potential confusers such as occlusions, dense traffic, and vehicles drifting away from the road. Aerial imagery typically yields a relatively small number of pixels on a target (roughly 20-100 pixels) and comparatively lower sampling rates ($1$-$2$ Hz) than common traditional rates ($25$-$60$ Hz) degrading the performance of apperance-based tracking-by-detection methods. Different sensor modalities such as infrared \cite{bhattacharya2011moving,gong2014joint,cao2015two}, \textit{Wide Area Motion Imagery (WAMI)}  \cite{teutsch2016robust,chen20153,cormier2016low} and
RGB \cite{tuermer2013airborne,zhang2013real,yousefhussien2016online,elliethy2016joint} have all shown the potential to improve tracking, however most of them perform poorly to achieve persistent tracking in real-time due to the unique challenges posed by aerial imagery or dependency on external sources of information (e.g. road map information) for achieving optimum results.  

For these reasons, the research community has started to explore multi-modal or adaptive sensor concepts. Uzkent, Hoffman and Vodacek \cite{uzkent2016real} use a multi-modal sensor concept - a wide field of view (FOV) panchromatic sensor coupled with a narrow FOV hyperspectral sensor to design a real-time persistent aerial tracking method that supersedes results observed in the above platforms. Hyperspectral sensors can provide unique spectral signatures for robustly distinguishing materials that can be used to solve surveillance tasks ranging from normal city traffic monitoring to military applications like border security. This can address some of the problems faced in infrared and single-band imaging such as discriminating foreground from background when the contrast is too low and tracking through traffic/clutter, e.g. Fig. \ref{fig:specprofiles} shows spectral signatures for background (building, trees, roads) as well as foreground objects which can be easily distinguished visually and hence, analytically as well. Along similar lines, if we can identify the signature of an upcoming potential occlusion (such as a patch of shadows), an adaptive sensing strategy could be employed to switch to a different modality such as polarization as shown in \cite{zhang2009polarimetric,zhang2009detecting}. Spectral signatures acquired by a hyperspectral camera (Fig. \ref{fig:specprofiles}) can be used to identify places of probable occlusion beforehand and account for them without the need for designing complex models and classifiers in grayscale and color imagery.

\begin{figure}[h]
\begin{center}
 \includegraphics[width=1\linewidth]{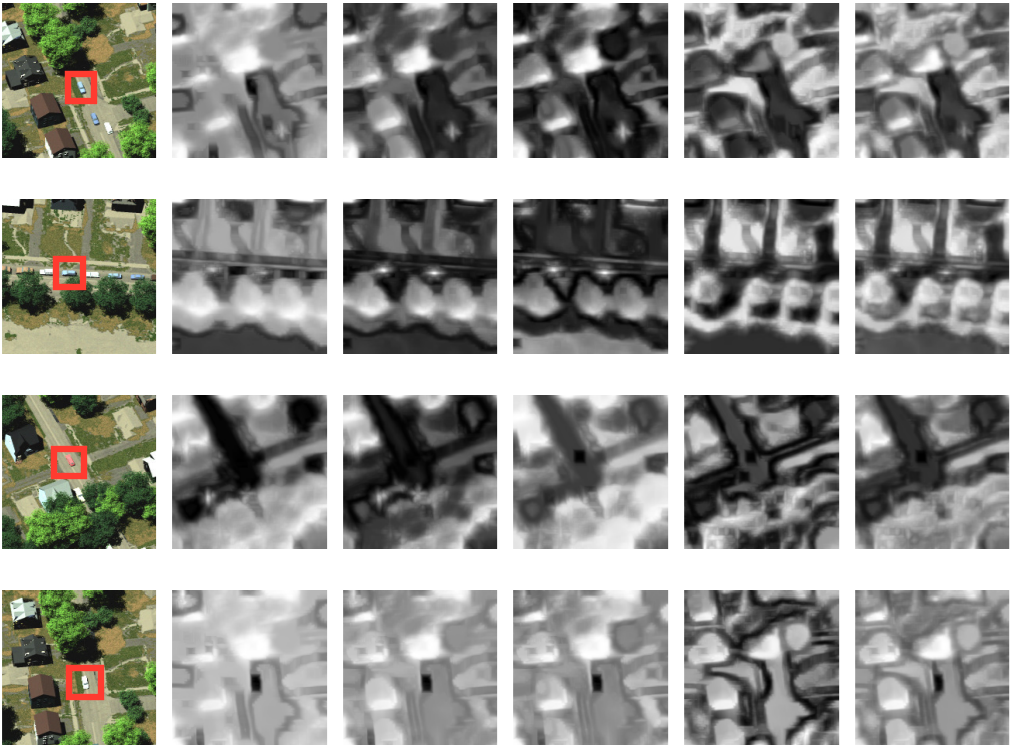}
 \\[-3ex]
\subfloat[]{\hspace{.166\linewidth}}
\subfloat[]{\hspace{.166\linewidth}}
\subfloat[]{\hspace{.166\linewidth}}
\subfloat[]{\hspace{.166\linewidth}}
\subfloat[]{\hspace{.166\linewidth}}
\subfloat[]{\hspace{.166\linewidth}}
\end{center}
   \caption{(a) shows the RGB ROI including the target of interest whereas (b,c,d) denote the likelihood maps from the blue, green, and red light, (e) represents the IR band and (f) represents the classic fusion (likelihood maps fused by equal weights). With the exception of the white car (last case), classic fusion actually makes triangulating the target of interest difficult rather than helpful.}
\label{fig:likelyhood1}
\end{figure}

A crucial part of any tracking framework is detecting the target of interest in a given region. Hyperspectral imagery generally has anything between 50 - 400 bands, which is a huge amount of data to be processed and transmitted. Conventionally, these large bands are reduced to a significantly smaller number of bands by either applying dimensionality reduction techniques (e.g. PCA) or averaging/sub-sampling (as the \cite{uzkent2016real}) before any processing is carried out. This may result in loss of valuable information that may be present in those dropped bands (as observed in Fig. \ref{fig:likelyhood1}). 

\begin{figure}[h]
\begin{center}
   \includegraphics[width=1\linewidth]{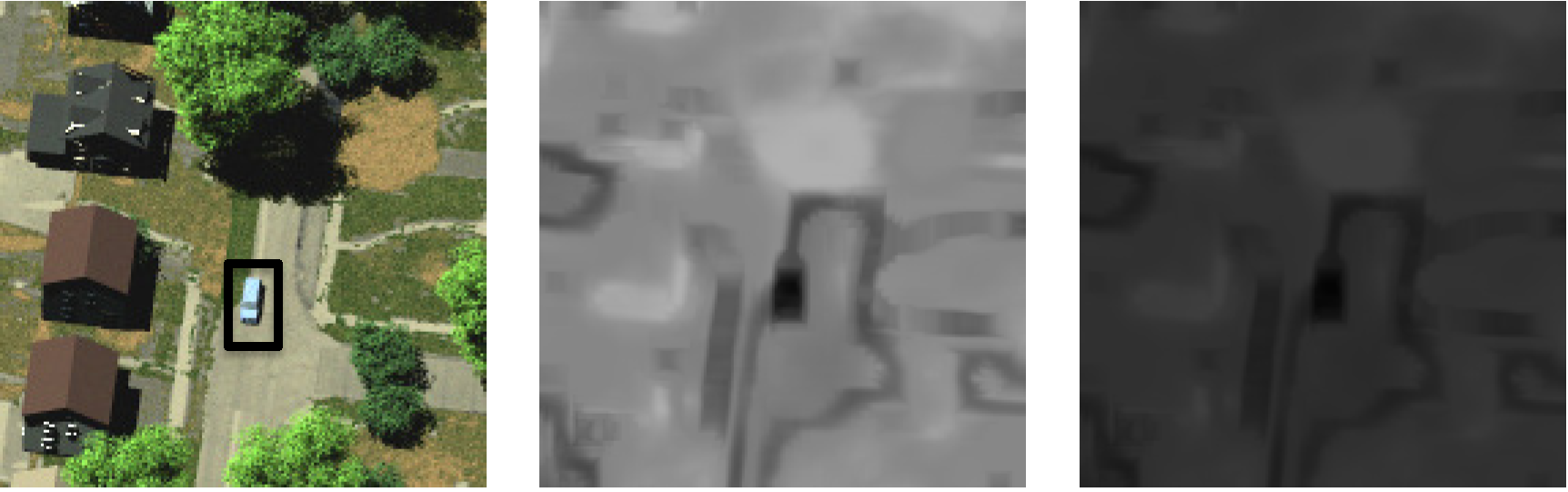}
   \\[-3ex]
   \subfloat[ROI]{\hspace{.33\linewidth}}
  \subfloat[Proposed Fusion]{\hspace{.33\linewidth}}
  \subfloat[Classic Fusion]{\hspace{.33\linewidth}}
\end{center}
   \caption{(a) shows the target of interest whereas (b) displays the result of the proposed hyperspectral likelihood maps fusion method and (c) represents the output of the classic fusion (equal weights to all bands). As seen, our method provides better contrast in the final likelihood map that helps triangulate the target of interest as compared to (c).}
\label{fig:wow1}
\end{figure}

In this paper, we propose the Hyperspectral Likelihoods-aided Tracker (HLT) with the following contributions: 

\begin{itemize}
\item An adaptive likelihood map fusion method that accounts for variability in the hyperspectral bands' data and calculates weights to fuse them such that valuable information from all bands is preserved. 
\item A novel target detection method that minimizes the user effort for training offline classifiers and tuning their hyperspectral parameters and dependency on other sources of information. 
\item A grouping method to group certain wavelengths channels together and represents them in a single likelihood map to achieve real-time target tracking.
\end{itemize}
A small preview of the result of our fusion method is shown in Fig. \ref{fig:wow1}.

In the following sections, we provide insight into the work done by the community towards aerial vehicle tracking and how hyperspectral imaging can help tackle the unique challenges posed by aerial platforms. Section \ref{rw} discusses very briefly the works done in different imaging domains. Section \ref{algo} and Section \ref{results} describe our algorithm and the results we obtained, and Section \ref{futurework} summarizes the work.

\section{Related Work} \label{rw}

Our work is inspired by Yin, Porikli and Collins \cite{yin2008likelihood} who proposed a single target tracker for aerial imagery and Palaniappan et al. \cite{palaniappan2010efficient} who used different set of visual features to extract feature likelihood maps and adaptively fuse them to obtain a single fusion map. While hyperspectral imagery has shown to have potential for vehicle tracking \cite{kerekes2006vehicle,uzkent2016integrating,uzkent2016real,uzkent2015feature,uzkent2013feature,uzkent2015spectral,uzkent2015efficient}, the
sensors are rare and expensive and hence most of the aerial tracking work has been done in infrared, single-band or RGB. In this section, we review the efforts done in the above sensor domains to solve vehicle detection and tracking and how hyperspectral imagery could possibly solve some of the problems. 

\subsection{Infrared Imagery}
Infrared imagery is helpful for object detection and tracking since it can pick up heat signatures emitted by the objects that conventional cameras are not capable of seeing. It has the ability to penetrate through smoke, fog and is generally insensitive to changes in light condition.  The COCOALIGHT system \cite{bhattacharya2011moving} proposed by Bhattacharya et al. consists of three phases - motion compensation, object detection and lastly, object tracking within and beyond the visible spectrum. The authors used feature-based (e.g. KLT \cite{shi1994good}) image registration to compensate for motion of the aerial imaging platform and then adopted cumulative frame differencing to detect and track foreground objects. Gong et al. \cite{gong2014joint} proposed ATR-Seg, a shape-aware manifold-learning based algorithm for joint tracking-recognition-segmentation in infrared imagery. Cao et al. \cite{cao2015two} present two new frameworks that use local clustering segmentation and kernel-based tracking theory to improve accuracy in target detection and tracking. However, since thermal imagery does not give unique fingerprints for different objects of the same category and is very contrast dependent, the above methods suffer during cluttered scenarios. 

\subsection{WAMI}

The WAMI sensor platform has received reasonable attention as it provides single band imagery covering a large area with higher spatial and temporal resolution than satellite imagery. It consists of camera array stitched together e.g. the widely used WPAFB 2009 \cite{WPAFB} dataset uses a matrix of 6 cameras to form a combined image of an overlooked scene. Moving object detection in WAMI is very challenging due to factors such as split and merged detections, weak contrast between object and background, shadows, and occlusions; most of which are already discussed above as potential obstacles towards successful aerial tracking (for survey of the algorithms used in WAMI, see \cite{sommer2016survey}). A recent approach by Teutsch and Grinberg \cite{teutsch2016robust} attempt to solve tracking vehicles in WAMI with promising results, however a large number of false negatives still occur during the initial 2-frame differencing for making object proposals. Chen and Medioni \cite{chen20153} proposed the use of 3-D mediated approaches to detect vehicles independent of parallax and registration errors. They then use a Bayesian network based data association method to link vehicle detection with their corresponding tracks. Cormier, Sommer and Teutsch \cite{cormier2016low} show that a combined descriptor of Local Binary Patterns (LBP) and Local Variance Measure
(VAR) histograms with the usage of Hellinger distance as a measure for similarity can significantly improve vehicle detection in WAMI data. Recently, Yi et al. \cite{yi2016vehicle} proposed the use of deep networks for classifying given patches as vehicles or non-vehicles in WAMI data. The fine-tuned AlexNet \cite{krizhevsky2012imagenet} outperms the HoG+Linear SVM classifier by $5\%$, however the fine-tuning and testing samples are collected on the same video, reducing the generalization capability of the network model. 

\subsection{RGB Aerial Imagery}

Detection, tracking and most recently, the counting of targets from satellite or aerial platform are useful for both commercial and government purposes. Tuermer et al. \cite{tuermer2013airborne} uses a disparity maps based approach with prior information of the road structure, orientation during image capture, and a global digital elevation model (DEM) to narrow down areas that can have vehicles. They then use an offline-trained histogram of oriented gradients (HoG) classifier for vehicle detection. Zhang et al. \cite{zhang2013real} proposed an online discriminative feature selection method that couples the classifier score with the
importance of samples, leading to a more robust and efficient
tracker. Yousefhussien, Browning and Kanan \cite{yousefhussien2016online} propose the Smooth Pursuit Tracking (SPT) algorithm which uses three kinds of saliency maps: appearance, location, and motion to track objects under all conditions, including long-term occlusion. Elliethy and Sharma \cite{elliethy2016joint} proposed an innovative approach to register captured WAMI frames with vector road map data(Open Street Map \cite{OpenStreetMap}) and track vehicles within those registered frames simultaneously, leading to efficient results. 
\begin{figure*}[ht]
\begin{center}
\includegraphics[width=0.9\linewidth]{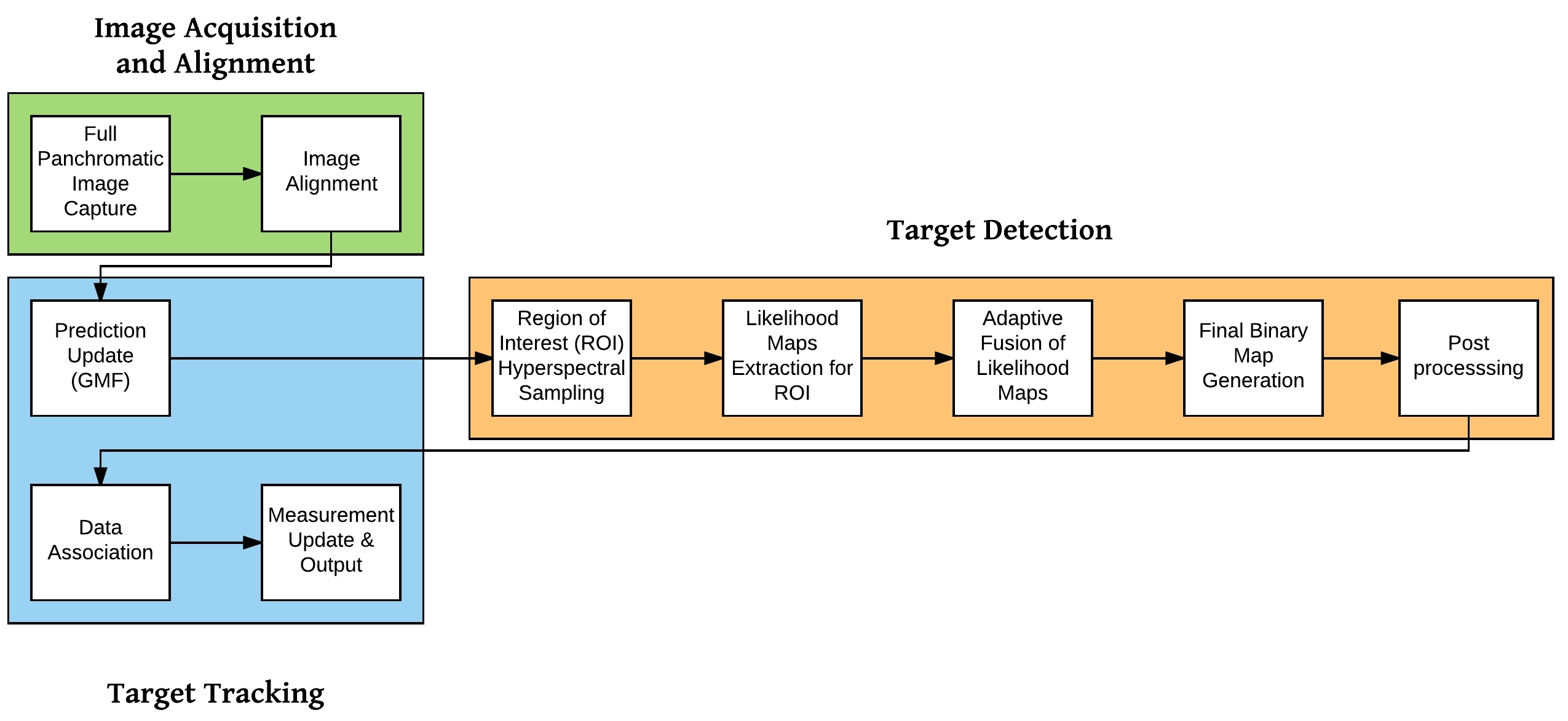}
\end{center}
   \caption{The proposed framework Hyperspectral Likelihoods-aided Tracker (HLT) showing major components: Image Acquisition and Alignment, Target Detection and Target Tracking. While we follow conventional methods for the first and the last part, our novelty lies in the target detection module.}
\label{fig:workflow}
\end{figure*}
\subsection{Spectral Aerial Imagery}

Hyperspectral imagery captures spectral signatures that can be used to uniquely characterize the materials in a given scene, the most common application being vegetation and mineral separation. Kerekes et al.\cite{kerekes2006vehicle} showed that it is possible to uniquely associate a vehicle in one image with its location in a subsequent image by using matched filter, given some constraints are imposed on the scene under consideration. The recent work of Svejkosky \cite{svejcars} shows that the spectral signatures of vehicles in hyperspectral imagery exhibit temporal variations due to changes in illumination, road surface properties and vehicle orientation - which justifies the outcomes in \cite{kerekes2006vehicle}. Uzkent et al. \cite{uzkent2016integrating} use single band WAMI-like imagery from a fixed platform to perform background subtraction for motion detection. Hyperspectral sensing is applied on the motion blobs to assign them hyperspectral likelihoods and filter hyperspectrally unlikely blobs. This is further improved by integrating the hyperspectral and kinematic likelihoods in a multi-dimensional assignment algorithm which better handles the changes in illumination and vehicle orientation specified in Kerekes et al. \cite{kerekes2006vehicle}. Such a framework can also handle mild occlusions as it includes the evolution of both kinematic and hyperspectral likelihoods. Uzkent et al. \cite{uzkent2016real} use a similar sensor on a moving platform to perform single target tracking. Different to \cite{uzkent2016integrating}, it follows the concept of tracking-by-detection methods by sampling a narrow FOV region of interest (ROI) hyperspectrally to search for the target. They avoid background subtraction in the moving platform case and instead use only narrow ROI hyperspectral image to get detection mask. The normalized difference vegetation index (NDVI) metric is used to detect vegetation dominated pixels using near-infrared and red wavelength bands. Cascaded to vegetation detection, a pixel-wise hyperspectral road classifier with non-linear SVM is applied to remove road pixels. These two classifiers are used to optimize the search space wherein HoG coupled with a linear SVM is used for removing non-vehicle blobs. The proposed approach shows promising results, however it relies on offline-trained classifiers which requires a considerable amount of training data collection and hyper-parameter tuning.

\section{Proposed Approach} \label{algo}

In the aforementioned papers, a lot of human effort is required in determining the best feature set and initializing parameters for model training since the performance of the framework depends on the former being carried out properly. To make things simpler, we propose the Hyperspectral Likelihoods-aided Tracker (HLT), an 'online' single target tracking framework on an aerial platform using a multi-modal sensor setup. 

An advantage of using hyperspectral imagery is that we have the freedom of leveraging information from multiple bands and picking the combination of bands that best helps our case. Our framework uses the fact that objects with certain spectral signatures can be better discriminated from the background in particular bands.  For example, a blue object is less reflective to blue light whereas it is relatively more reflective to red, and green light. On the other hand, an object of class 'vegetation' absorbs the blue and red light more than the green light as seen in Fig. \ref{fig:specprofiles}, causing the plant to look greener. By concatenating individual bands’ histograms without using this information can result in likelihood maps with less inter-class variance.

HLT's work flow can be divided into three parts: Image Acquisition and Camera Motion Removal (Alignment), Target Detection and Target Tracking, which can be visualized in Fig. \ref{fig:workflow}. While we follow conventional methods for the first and the last part, the novelty of our algorithm lies in how we process the target detection module at near real-time speeds.

\subsection{Image Acquisition and Alignment}

As part of a multi-modal sensor framework, we have two images at our hands for each frame: a panchromatic wide FOV image and a hyperspectral (400 nm - 1000 nm; 61 bands) narrow FOV image. For our convenience, we drop the last (1000 nm) band and hence are left with 60 bands at our disposal. In this section, we use consecutive panchromatic images to compute the homographies between two successive frames using SIFT \cite{lowe2004distinctive} and RANSAC \cite{fischler1981random}. This step is required as we perform multi-dimensional (frames) blob-to-track association method to correct the false assignments in light of later observations. Additionally, the large camera motion in between two successive frames can lead to false prediction without alignment. With alignment and better filtering, we can keep the hyperspectral region of interest (ROI) as narrow as possible, thus speeding up the frame rate of the system.

\subsection{Target Detection} \label{TD}

Initially, the predicted bounding box area is obtained from the tracker and sampled hyperspectrally, which results in a 200x200 pixels image of each hyperspectral band. Next, we generate a likelihood map for each band image as follows: We take all the 60 bands at our disposal. To reduce the computational load, we exploit the fact that neighboring bands contribute to the final likelihood map similarly. In this direction, we propose a grouping based approach to use the correlation in between the same group wavelength maps. To do so, we consider computing two likelihood maps in every 100 nm: in between 400 and 500 nm, two likelihood maps are computed by grouping the histograms in between 400 and 450 nm and 460 and 500 nm. We divide our 60 bands into groups of 12 and use the integral image theorem to obtain a 10-bin histogram representation for each band - leading to each group's 50-D feature vector representation. All histograms are normalized to improve robustness against lighting changes. We then use three sliding windows: 20 x 10, 10 x 20 and 14 x 14 to compare the spectral pdf and the target model's pdf using $\chi^2$ distance metric and obtain 3 potential likelihood maps. The one with the highest confidence is picked out of three pixel-wise confidences. The three sliding windows relate to an estimate of the pixels occupied by a target in the sampled window - going vertical, going horizontal and moving in diagonal manner with the same scale considering the constant altitude during flight time.

Now that we have 12 likelihood maps amongst all 60 bands in consideration, there is still a need to fuse them and generate one final likelihood map. Techniques such as Variance Ratio method \cite{yin2008likelihood} and Sum-rule method \cite{kittler1998weighted} (discussed briefly in Section \ref{results}) do not take into consideration spectral correlation. Here, we adaptively fuse the maps in the following manner: We obtain a specific threshold $T_{i}=\lbrace i=1,2,3,...\rbrace$ for every likelihood map $L_{i}=\lbrace i=1,2,3,...\rbrace$ using multilevel Otsu's thresholding method as described in \cite{uzkent2016real}. Once the thresholds are estimated, we apply them to the likelihood maps to estimate corresponding binary maps as shown in Eq. \ref{eq:1}:
\renewcommand{\arraystretch}{1.0}
\begin{equation} \label{eq:1}
B_{i}(x,y) = \left\{
  \begin{array}{l l}
    0 \quad if\:L_{i}(x,y)>T_{i} \\
    1 \quad otherwise
  \end{array}
\right.,
\end{equation}
After computing the binary maps, the coefficients are estimated by considering the positive pixels ($1s$) as
\begin{equation} \label{eq:2}
	c_{i} = \dfrac{sum(B_{i})}{\sum_{j=1}^{N}sum(B_{j})}.
\end{equation}
where $N$ represents the number of bands. From Eq. \ref{eq:2}, the $L_{i}$ with large number of positive pixels in $B_{i}$ are assigned large values. However, we want to assign smaller weights to these likelihood maps since they contain many false positives i.e. the contribution of that band group is small. To do so, we convert the coefficients to meaningful values with logistic function shown as:
\begin{equation}
	w_{i} = \dfrac{1}{1+exp^{-k(c_{i}-x_{0})}}
\label{eq:logistic_function}
\end{equation}
where $k$ and $x_{0}$ are preset parameters. We normalize the weights with $\textit{L-1}$ normalization to have unit weight vector as:
\begin{equation}
	w_{i} = \dfrac{w_{i}}{\norm{w}_{1}}.
\end{equation}
The final likelihood map fusion is formulated as follows:
\begin{equation} \label{eq:f}
	L_{Final} = \sum_{i}^{N}w_{i}L_{i}.
\end{equation}
For this step, since fusing $N$ number of likelihood maps (in this case, 60) would have been computationally expensive, we exploited the fact that neighboring bands contribute to the final likelihood map similarly and grouped them together, helping us achieve near real-time computations. The fusion also ensures that data from all bands is effectively taken into consideration, as opposed by conventional techniques that use dimensionality reduction algorithms (e.g. PCA) to obtain fewer bands. 
\begin{figure}[t]
\begin{center}
   \includegraphics[width=1\linewidth]{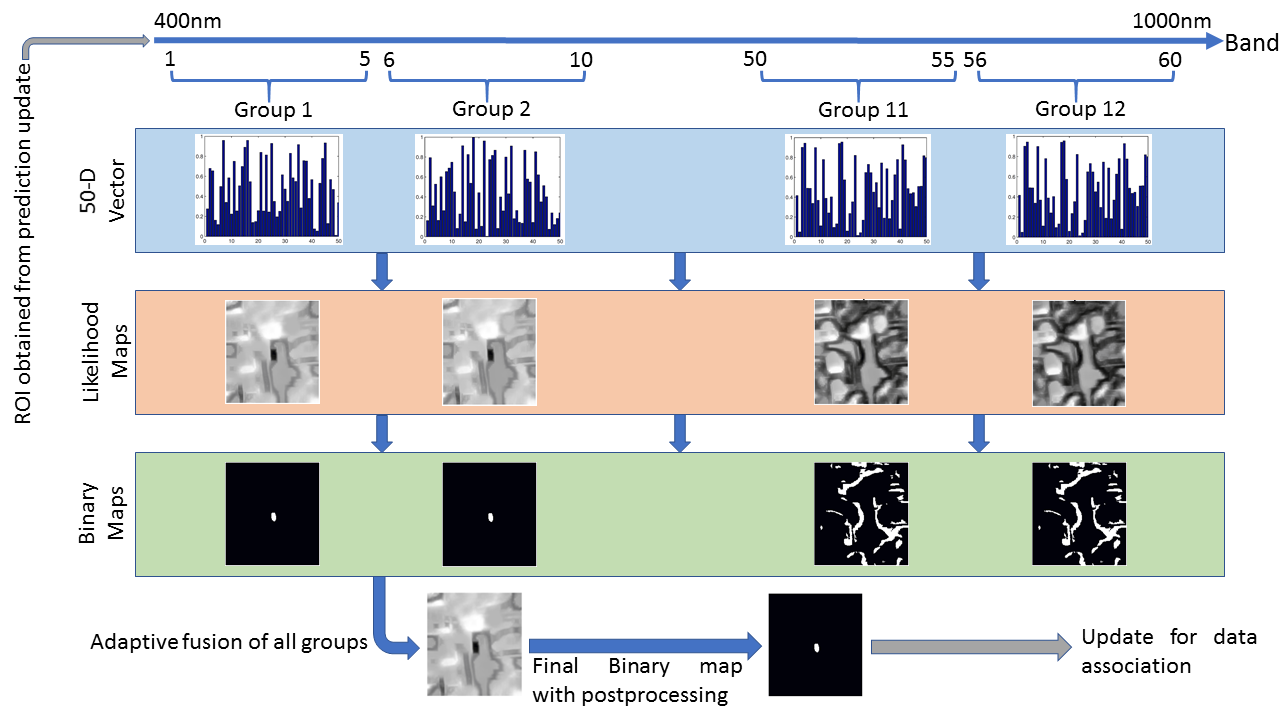}
\end{center}
   \caption{The grouping framework discussed in the Target Detection (\ref{TD}) section. The likelihood maps of all 12 groups are thresholded to  binary maps which are adaptively processed to obtain weighting parameters for fusing the groups. This leads to a final likelihood and binary map that marks the possible location(s) of the target-of-interest in the scene.}
\label{fig:groupingframework}
\end{figure}
Eq. \ref{eq:f} gives us the final likelihood map which is again thresholded to obtain the final binary map. Morphological closing and connected component labeling post-processing methods are then performed to determine the final set of candidate blobs which is passed to the tracking part of the algorithm. This entire flow is summarized in Fig. \ref{fig:groupingframework}. The optimal candidate blob is then determined by the target tracking framework discussed below.

One disadvantage of this method is it does not handle tracking robustly when the scene is cluttered with objects having similar paint models to the target. The proposed framework will try to assign smaller weights to the discriminative maps as the number of positive pixels will be large as expected. However, we consider the evolution of the target's motion in our data association framework complementing the proposed detection method in these cases.

\subsection{Target Tracking} \label{TT}

The high-level description of the tracking framework highlights the fact that we do not use the full-frame panchromatic imagery in the tracking task. The homographies computed in the registration module are used to update the track statistics on a canonical plane. We follow the Mixture of Extended and Linear Kalman Filter Banks approach similar to \cite{uzkent2016real,uzkent2016integrating}. By using multiple filters, we can accommodate different motion models which approximate different scenarios such as turning right, left and going straight. The Extended Kalman Filter leads to better handling of mildly non-linear motion density functions. This filtering framework, coupled with Gaussian Mixture Filter (GMF), helps us sample a ROI that is more likely to contain the target in an incoming frame. After the detection step (see~\ref{TD}), the detected blobs are passed to the Multi-dimensional Hyperspectral and Motion Likelihood Aided Assignment algorithm (covered extensively in \cite{uzkent2015spectral}). Hyperspectral likelihoods are computed for each blob by comparing the spectral histogram features of the blob and reference model using the $\chi^2$ distance metric. Finally, the track statistics (i.e. the track's state space matrix) are updated with the new set of measurements from the previous $N$ scans.

\section{Experiments and Results} \label{results}

Our synthetic and real hyperspectral videos are described in the section below followed by the experiments and results. 

\subsection{Data}

\subsubsection{Synthetic DIRSIG Data}
\begin{figure}[h]
\begin{center}
   \includegraphics[width=1\linewidth]{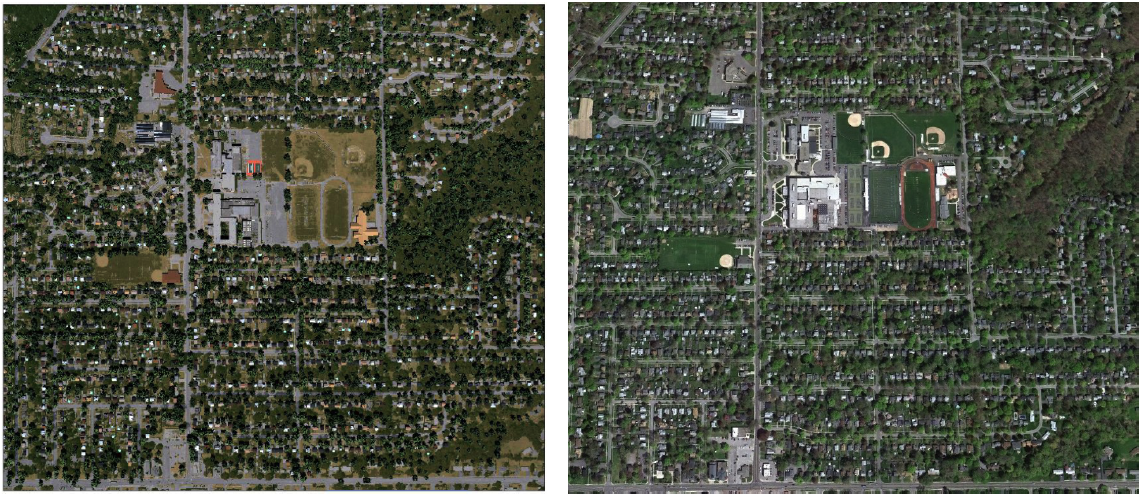}
   \\[-3ex]
   \subfloat[DIRSIG]{\hspace{.5\linewidth}}
   \subfloat[Google Maps]{\hspace{.5\linewidth}}
\end{center}
   \caption{(a) DIRSIG Nadir RGB image of Megascene \#1 area in northern Rochester, NY and (b) same region as observed via Google Maps.}
\label{fig:DIRSScenes}
\end{figure}

We use synthetic imagery generated by the \textit{Digital Imaging and Remote Sensing Image Generation} (DIRSIG) model to develop and test the approach in a user-controlled environment that also provides us solid ground truth without the need for annotation.  DIRSIG \cite{schott1999advanced,ientilucci2003advances} is capable of and has been proved to produce imagery in a variety of modalities, such as multispectral, hyperspectral, polarimetric, LIDAR, etc. in the visible through thermal infrared regions of the electromagnetic (EM) spectrum. Our motivation for using DIRSIG's synthetic data is two fold:
\begin{itemize}
\item We know the true positions of all objects in a synthetic image and can accurately compute performance metrics for the tracking system.
\item The use of multi-modal sensors in Unmanned Aerial Vehicles (UAV) is a new and evolving area of research. Costs associated with procuring hardware, such as hyperspectral imagery devices is fairly expensive and cost prohibitive.
\end{itemize}
The scenario used in this paper comes from the DIRSIG Megascene \#1 area, a high fidelity artificial recreation of natural and man-made objects in a vast region of the northern Rochester, NY (Fig. \ref{fig:DIRSScenes}).

We simulated the flight of the platform across the entire scene shown in Fig. \ref{fig:DIRSScenes} with 90 m/s constant velocity at an altitude of 3000 m under varying atmospheric and time conditions (comparable to the simulation carried out in \cite{uzkent2016real}). The spectral range was 400 to 1000 nm with a spectral resolution of 10 nm and the average ground sampling distance (GSD) was set to 0.30 m. We used Simulation of Urban Mobility (SUMO) \cite{krajzewicz2002sumo}, integrated with DIRSIG, to produce dynamic images with vehicles moving along a predefined path. A total of 88 vehicles were placed in scenes that generated 130 seconds of video each.  Vehicle movement was modeled by a normal distribution $N(\mu, \sigma)$, where $\mu$ was set to 35 mph (i.e., the average speed limit in the part of Rochester, NY under consideration) and $\sigma$ was 10. Finally, we assign 24 different paints with equal probability to induce extra variance in the vehicles generated. To simulate a challenging environment, we also incorporate traffic lights in one of the intersections - thus modeling a nearly perfect urban scene, the only exception being the presence of pedestrians. Overall, we obtain two images per frame - a wide FOV panchromatic imagery that is used for image registration and a narrow FOV hyperspectral imagery that is used for vehicle detection and tracking. Since most materials can be identified by their reflectance spectra (Fig. \ref{fig:specprofiles}), we convert all hyperspectral images from radiance to reflectance before any processing is carried out. 

\subsubsection{University Parking Lot Data}

To capture real data, we used a hyperspectral camera that collects 30 channels in the visible wavelength range at 2 Hz. The scenes were captured overlooking an university's parking lot at a non-nadir point of view. The hyperspectral image obtained was of the size 184 x 200 pixels, similar to the size of the ROI given to the target detection method.

\subsection{Experimental Results}

We test the adaptive likelihood map fusion approach in the scenario described in \cite{uzkent2016real} to compare our results. We select 43 of the 88 rendered cars and initialize the first frame with a bounding box selection representative of the target of interest. We consider the \textit{Track Purity} (TrP) and \textit{Target Purity} (TgP) metrics to measure tracking performance. TrP evaluates the ratio of the number of frames a dominant target is assigned to the track to the track life. On the other hand, TgP takes into account the ground truth life. TrP favors short tracks (terminated) whereas the TgP can provide more fair evaluation in cases where the track life is shorter than ground truth life.

We compare our proposed adaptive fusion approach to several cases and kinematic data based trackers across various sensor modalities as seen in Table \ref{table: Comparisons}. We have already discussed online discriminative feature selection learning (OFDS) \cite{zhang2013real} and Hyperspectral Feature based Tracker (HFT) \cite{uzkent2016real} (Note: \cite{uzkent2016real} also uses the same synthetic data). Additionally, we introduce a recent state-of-the-art aerial vehicle tracker, LoFT (wide-area aerial tracker via likelihood of features tracking) \cite{pelapur2012persistent}, an improved version of \cite{palaniappan2010efficient} for WAMI data - it should be noted that the source code for LoFT is not publicly available and hence we use the results from their paper itself since the GSD's and temporal resolution for datasets used in the papers is very similar. We also compare our results to two popular band fusion methods: the Variance Ratio method \cite{yin2008likelihood} and Sum-rule method  \cite{kittler1998weighted}. Variance Ratio method considers uncorrelated likelihood maps - namely Intensity Histogram, HoG, Motion History, Saliency and Template likelihood maps, and selects weights such that the map that maximizes the separability between the foreground target and its surrounding background region is given higher importance. On similar lines, Sum-rule method concatenates individual likelihood maps with equal weights and fuses them. We also compare the adaptive fusion with hyperspectral data to tracking with grayscale, RGB and multispectral (MS) data. For RGB, the bands are sampled in the central wavelength of Red, Green and Blue light with the same spectral resolution (10 nm). In multispectral case, 6 bands are considered to represent every pixel in a ROI. These bands are sampled in the full spectrum range. 

\begin{table}[tbp]
\centering

\begin{tabular}{|l|l|l|}
\hline
\textbf{Tracker}  & \textbf{Track Purity}   & \textbf{Target Purity}  \\ \hline
 \hline
Gray-scale Data & 28.43          & 04.28          \\ \hline
RGB Data        & 39.20          & 35.07          \\ \hline
MS Data         & 55.12          & 50.91          \\ \hline
OFDS \cite{zhang2013real}           & 12.66          & 12.66          \\ \hline
LoFT \cite{pelapur2012persistent}           & 60.30          & 40.50          \\ \hline
Sum-rule \cite{kittler1998weighted}       & 50.17          & 45.25          \\ \hline
Variance Ratio \cite{yin2008likelihood} & 48.26          & 44.56          \\ \hline
HFT \cite{uzkent2016real}  & 69.78 		 & 60.30 \\ \hline
Ours (HLT)      & 64.37          & 57.49          \\ \hline
\end{tabular}
\\

\caption{Comparison of the proposed HLT tracker with other trackers.}
\label{table: Comparisons}
\end{table}

As seen in Table \ref{table: Comparisons}, the proposed adaptive fusion approach outperforms the current fusion methods: Sum-rule and Variance Ratio based likelihood map fusion approach by a large margin. Our approach is a global method that can better determine the usefulness of the distance map by taking account the label of each pixel in a ROI. The tracking-by-detection algorithms (OFDS) perform poorly as this is a low temporal and spatial resolution scenario. As expected, the proposed fusion on hyperspectral data outperforms gray-scale, RGB and multispectral data as they contain less information on the objects. The proposed fusion outperforms the other fusion methods by a large margin. Finally, the proposed fusion method is outperformed slightly by HFT \cite{uzkent2016real}, however, the HFT tracker requires designing and training classifiers offline together with reasonable number of parameter tuning for the vegetation detection, pixel-wise road classifier and vehicle detector. On the other hand, our method does not use any offline-trained classifier and requires only two parameter tuning (see eq.~\ref{eq:logistic_function}) in the target detection part.

We also test the adaptive fusion approach with different band grouping strategies. In our original approach, we used 12 groups out of the 60 bands - every group is represented by 50-D feature vector from 5 adjacent bands. The results for different grouping strategies are shown in Table \ref{table: Groups}. This concludes that using 12 groups was the best option amongst all grouping methods.

\begin{table}[h]
\centering
\small
\begin{tabular}{|l|l|l|}
\hline
\textbf{Tracker}  & \textbf{Track Purity}   & \textbf{Target Purity}  \\ \hline
 \hline
2 Groups       & 48.50          & 45.28          \\ \hline
3 Groups       & 55.27          & 48.81          \\ \hline
6 Groups       & 60.11          & 55.39          \\ \hline
12 Groups      & 64.37          & 57.49          \\ \hline
20 Groups 	   & 55.65          & 51.50          \\ \hline
\end{tabular}
\\
\caption{The results of the grouping by using different number of neighboring bands are shown in the table. Twelve groups are shown to be the optimum number of groups for $60$ bands in the 400-1000 nm wavelength range. Twenty groups (3 neighboring bands) shows inferior performance as 3 neighbors might result in less discriminative histogram features than the 5 neighbors case. Also, 6 groups (10 neighboring bands) is outperformed by 12 groups as the groups contain first five and second five bands where the contribution of the first half can be reduced by the less discriminative second half. We believe that each 50 nm (400-450 nm, 450-500 nm, ... 950-1000 nm) contains more useful histogram features than any other combination resulting in more discriminative final fusion likelihood map.}
\label{table: Groups}
\end{table}

\begin{table}[h]
\small
\centering
\begin{tabular}{|l|l|l|}
\hline
\textbf{Module}  & \textbf{Run-time}  & \textbf{Allocated Time}\\ \hline
\hline
Gradient Integral Images   & 0.21  s.& - \\ \hline
Likelihood Map Fusion      & 0.16 s. & - \\ \hline
Detection Module       	   & 0.37  s.& 0.35 s.\\ \hline
\end{tabular}
\\
\caption{Run time performances of the detection modules. Experiments were carried out on a Linux server with AMD Opteron 4226/2.7 GHz processor. We code the detection modules in C and couple them to MATLAB platform with the mex compiler.}
\label{runtimedetection_2}
\end{table}

Table~\ref{runtimedetection_2} displays the run-time performances of the detection module which is the key contribution of this tracker. The proposed detection module of new HLT tracker matches the time (0.35 s.) allocated for the detection module. We want to highlight that the other parts of the HLT tracker are similar to the HFT tracker \cite{uzkent2016real}. To summarize the workflow, the HLT performs tracking at 1.43Hz as the HFT tracker. It proposes a new detection module running at the same operation rate to the detection module of the HFT tracker while minimizing the user effort that has to be performed offline.

\subsubsection{Preliminary Results on University Parking Lot Data}

\begin{figure}[h]
\begin{center}
   \includegraphics[width=1\linewidth]{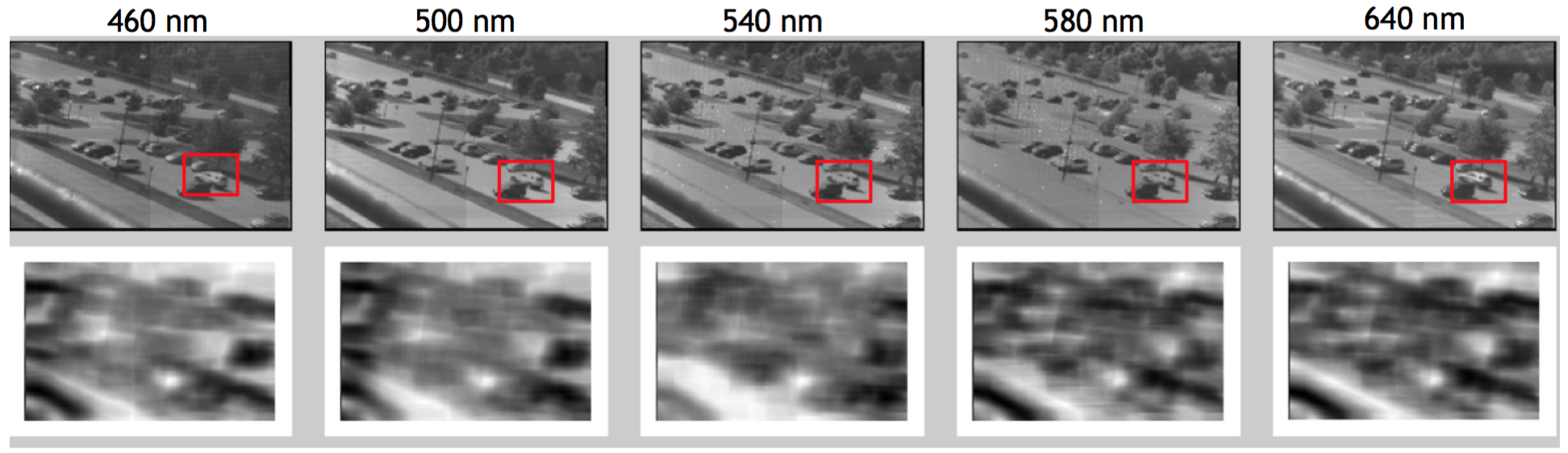}
\end{center}
   \caption{The object being tracked in the university's parking lot. Top row figures demonstrate some of the band imagery whereas bottom row figures show corresponding likelihood maps.}
\label{fig:realvisualresults}
\end{figure}

We increase the validity of the proposed approach by performing a high-level preliminary test on real data captured at a parking lot of an university in Rochester, NY. The size of the frames is 184 x 200 pixels which is similar to the ROI sampled hyperspectrally in the synthetic dataset. Fig. \ref{fig:realvisualresults} shows some channels and corresponding hyperspectral likelihood maps extracted in the same way as in the section~\ref{TD}. It is seen that blue wavelength describes our target better than the others. If classic fusion or dimensionality reduction was applied, this particular band information would probably be lost. Since our method works by adaptively assigning weights, one can assume that in this case, blue wavelength would have major importance during fusion.

\section{Conclusion} \label{futurework}
In this study, we propose a novel hyperspectral likelihood maps-aided tracking method inspired by an adaptive hyperspectral sensor. Our target detection method removes the need to build  offline classifiers and learns the generative target model in an online manner for hyperspectral channels. The foreground background separation methods are excluded resulting in a detection module more robust to the 3D parallax effect. The run-time of the detection part is optimized by finding the optimal subset of neighboring hyperspectral channels with high correlation to reduce number of likelihood maps we process. As an extension of this work, we plan on designing a framework that can adaptively group different hyperspectral channels to generate more discriminative likelihood maps. As flying UAVs with fitted sensors become more practical, we will capture real data and conduct more comprehensive tests to strengthen our claim on the currently published results. Also, we plan on building a large aerial hyperspectral dataset covering different scenarios to train a CNN to perform semantic vehicle detection and release this dataset to the research community.


{\small
\bibliographystyle{ieee}
\bibliography{egbib}

\begin{thebibliography}{10}\itemsep=-1pt

\bibitem{WPAFB}
AFRL.
\newblock Wright-patterson air force base (wpafb) dataset.
\newblock \url{https://www.sdms.afrl.af.mil/index.php?collection=wpafb2009},
  2009.

\bibitem{bhattacharya2011moving}
S.~Bhattacharya, H.~Idrees, I.~Saleemi, S.~Ali, and M.~Shah.
\newblock Moving object detection and tracking in forward looking infra-red
  aerial imagery.
\newblock In {\em Machine vision beyond visible spectrum}, pages 221--252.
  Springer, 2011.

\bibitem{cao2015two}
Y.~Cao, G.~Wang, D.~Yan, and Z.~Zhao.
\newblock Two algorithms for the detection and tracking of moving vehicle
  targets in aerial infrared image sequences.
\newblock {\em Remote Sensing}, 8(1):28, 2015.

\bibitem{chen20153}
B.-J. Chen and G.~Medioni.
\newblock 3-d mediated detection and tracking in wide area aerial surveillance.
\newblock In {\em Applications of Computer Vision (WACV), 2015 IEEE Winter
  Conference on}, pages 396--403. IEEE, 2015.

\bibitem{cormier2016low}
M.~Cormier, L.~W. Sommer, and M.~Teutsch.
\newblock Low resolution vehicle re-identification based on appearance features
  for wide area motion imagery.
\newblock In {\em 2016 IEEE Winter Applications of Computer Vision Workshops
  (WACVW)}, pages 1--7. IEEE, 2016.

\bibitem{elliethy2016joint}
A.~Elliethy and G.~Sharma.
\newblock A joint approach to vector road map registration and vehicle tracking
  for wide area motion imagery.
\newblock In {\em Acoustics, Speech and Signal Processing (ICASSP), 2016 IEEE
  International Conference on}, pages 1100--1104. IEEE, 2016.

\bibitem{fischler1981random}
M.~A. Fischler and R.~C. Bolles.
\newblock Random sample consensus: a paradigm for model fitting with
  applications to image analysis and automated cartography.
\newblock {\em Communications of the ACM}, 24(6):381--395, 1981.

\bibitem{gong2014joint}
J.~Gong, G.~Fan, L.~Yu, J.~P. Havlicek, D.~Chen, and N.~Fan.
\newblock Joint target tracking, recognition and segmentation for infrared
  imagery using a shape manifold-based level set.
\newblock {\em Sensors}, 14(6):10124--10145, 2014.

\bibitem{ientilucci2003advances}
E.~J. Ientilucci and S.~D. Brown.
\newblock Advances in wide-area hyperspectral image simulation.
\newblock In {\em AeroSense 2003}, pages 110--121. International Society for
  Optics and Photonics, 2003.

\bibitem{kerekes2006vehicle}
J.~Kerekes, M.~Muldowney, K.~Strackerjan, L.~Smith, and B.~Leahy.
\newblock Vehicle tracking with multi-temporal hyperspectral imagery.
\newblock In {\em Defense and Security Symposium}, pages 62330C--62330C.
  International Society for Optics and Photonics, 2006.

\bibitem{kittler1998weighted}
J.~Kittler and S.~Hojjatoleslami.
\newblock A weighted combination of classifiers employing shared and distinct
  representations.
\newblock In {\em Computer Vision and Pattern Recognition, 1998. Proceedings.
  1998 IEEE Computer Society Conference on}, pages 924--929. IEEE, 1998.

\bibitem{krajzewicz2002sumo}
D.~Krajzewicz, G.~Hertkorn, C.~R{\"o}ssel, and P.~Wagner.
\newblock {SUMO (Simulation of Urban MObility)-an open-source traffic
  simulation}.
\newblock In {\em Proceedings of the 4th middle East Symposium on Simulation
  and Modelling (MESM20002)}, pages 183--187, 2002.

\bibitem{krizhevsky2012imagenet}
A.~Krizhevsky, I.~Sutskever, and G.~E. Hinton.
\newblock Imagenet classification with deep convolutional neural networks.
\newblock In {\em Advances in neural information processing systems}, pages
  1097--1105, 2012.

\bibitem{lowe2004distinctive}
D.~G. Lowe.
\newblock Distinctive image features from scale-invariant keypoints.
\newblock {\em International journal of computer vision}, 60(2):91--110, 2004.

\bibitem{OpenStreetMap}
{OpenStreetMap contributors}.
\newblock {Planet dump retrieved from http://planet.osm.org }.
\newblock \url{http://www.openstreetmap.org}, 2017.

\bibitem{palaniappan2010efficient}
K.~Palaniappan, F.~Bunyak, P.~Kumar, I.~Ersoy, S.~Jaeger, K.~Ganguli,
  A.~Haridas, J.~Fraser, R.~M. Rao, and G.~Seetharaman.
\newblock Efficient feature extraction and likelihood fusion for vehicle
  tracking in low frame rate airborne video.
\newblock In {\em Information fusion (FUSION), 2010 13th Conference on}, pages
  1--8. IEEE, 2010.

\bibitem{pelapur2012persistent}
R.~Pelapur, S.~Candemir, F.~Bunyak, M.~Poostchi, G.~Seetharaman, and
  K.~Palaniappan.
\newblock Persistent target tracking using likelihood fusion in wide-area and
  full motion video sequences.
\newblock In {\em Information Fusion (FUSION), 2012 15th International
  Conference on}, pages 2420--2427. IEEE, 2012.

\bibitem{schott1999advanced}
J.~R. Schott, S.~D. Brown, R.~V. Raqueno, H.~N. Gross, and G.~Robinson.
\newblock An advanced synthetic image generation model and its application to
  multi/hyperspectral algorithm development.
\newblock {\em Canadian Journal of Remote Sensing}, 25(2):99--111, 1999.

\bibitem{shi1994good}
J.~Shi et~al.
\newblock Good features to track.
\newblock In {\em Computer Vision and Pattern Recognition, 1994. Proceedings
  CVPR'94., 1994 IEEE Computer Society Conference on}, pages 593--600. IEEE,
  1994.

\bibitem{sommer2016survey}
L.~W. Sommer, M.~Teutsch, T.~Schuchert, and J.~Beyerer.
\newblock A survey on moving object detection for wide area motion imagery.
\newblock In {\em Applications of Computer Vision (WACV), 2016 IEEE Winter
  Conference on}, pages 1--9. IEEE, 2016.

\bibitem{svejcars}
{Svejkosky, Joseph}.
\newblock {Hyperspectral Vehicle BRDF Learning: An Exploration of Vehicle
  Reflectance Variation and Optimal Measures of Spectral Similarity for Vehicle
  Reacquisition and Tracking Algorithms}.
\newblock \url{http://scholarworks.rit.edu/theses/9203}, 2016.

\bibitem{teutsch2016robust}
M.~Teutsch and M.~Grinberg.
\newblock Robust detection of moving vehicles in wide area motion imagery.
\newblock In {\em Proceedings of the IEEE Conference on Computer Vision and
  Pattern Recognition Workshops}, pages 27--35, 2016.

\bibitem{tuermer2013airborne}
S.~Tuermer, F.~Kurz, P.~Reinartz, and U.~Stilla.
\newblock Airborne vehicle detection in dense urban areas using {HoG} features
  and disparity maps.
\newblock {\em IEEE Journal of Selected Topics in Applied Earth Observations
  and Remote Sensing}, 6(6):2327--2337, 2013.

\bibitem{uzkent2015feature}
B.~Uzkent, M.~Hoffman, A.~Vodacek, and B.~Chen.
\newblock Feature matching with an adaptive optical sensor in a ground target
  tracking system.
\newblock {\em IEEE Sensors Journal}, 15(1):510--519, 2015.

\bibitem{uzkent2015efficient}
B.~Uzkent, M.~J. Hoffman, and A.~Vodacek.
\newblock Efficient integration of spectral features for vehicle tracking
  utilizing an adaptive sensor.
\newblock In {\em SPIE/IS\&T Electronic Imaging}, pages 940707--940707.
  International Society for Optics and Photonics, 2015.

\bibitem{uzkent2015spectral}
B.~Uzkent, M.~J. Hoffman, and A.~Vodacek.
\newblock Spectral validation of measurements in a vehicle tracking {DDDAS}.
\newblock {\em Procedia Computer Science}, 51:2493--2502, 2015.

\bibitem{uzkent2016integrating}
B.~Uzkent, M.~J. Hoffman, and A.~Vodacek.
\newblock {Integrating Hyperspectral Likelihoods in a Multidimensional
  Assignment Algorithm for Aerial Vehicle Tracking}.
\newblock {\em IEEE Journal of Selected Topics in Applied Earth Observations
  and Remote Sensing}, 9(9):4325--4333, 2016.

\bibitem{uzkent2016real}
B.~Uzkent, M.~J. Hoffman, and A.~Vodacek.
\newblock Real-time vehicle tracking in aerial video using hyperspectral
  features.
\newblock In {\em Proceedings of the IEEE Conference on Computer Vision and
  Pattern Recognition Workshops}, pages 36--44, 2016.

\bibitem{uzkent2013feature}
B.~Uzkent, M.~J. Hoffman, A.~Vodacek, J.~P. Kerekes, and B.~Chen.
\newblock Feature matching and adaptive prediction models in an object tracking
  {DDDAS}.
\newblock {\em Procedia Computer Science}, 18:1939--1948, 2013.

\bibitem{yi2016vehicle}
M.~Yi, F.~Yang, E.~Blasch, C.~Sheaff, K.~Liu, G.~Chen, and H.~Ling.
\newblock Vehicle classification in wami imagery using deep network.
\newblock In {\em SPIE Defense+ Security}, pages 98380E--98380E. International
  Society for Optics and Photonics, 2016.

\bibitem{yin2008likelihood}
Z.~Yin, F.~Porikli, and R.~T. Collins.
\newblock Likelihood map fusion for visual object tracking.
\newblock In {\em Applications of Computer Vision, 2008. WACV 2008. IEEE
  Workshop on}, pages 1--7. IEEE, 2008.

\bibitem{yousefhussien2016online}
M.~A. Yousefhussien, N.~A. Browning, and C.~Kanan.
\newblock Online tracking using saliency.
\newblock In {\em Applications of Computer Vision (WACV), 2016 IEEE Winter
  Conference on}, pages 1--10. IEEE, 2016.

\bibitem{zhang2009polarimetric}
C.~Zhang, H.~Cheng, Z.~Chen, W.~Zheng, and Y.~Cao.
\newblock Polarimetric imaging of camouflage screen in visible and infrared
  wave band [j].
\newblock {\em Infrared and Laser Engineering}, 38(3):424--427, 2009.

\bibitem{zhang2009detecting}
C.-Y. Zhang, H.-F. Cheng, Z.-H. Chen, and W.-W. ZHENG.
\newblock Detecting low reflectivity camouflage net by using polarization
  remote sensing.
\newblock {\em Journal of Infrared and Millimeter Waves}, 2:014, 2009.

\bibitem{zhang2013real}
K.~Zhang, L.~Zhang, and M.-H. Yang.
\newblock Real-time object tracking via online discriminative feature
  selection.
\newblock {\em IEEE Transactions on Image Processing}, 22(12):4664--4677, 2013.

\end{thebibliography}
}

\end{document}